\documentclass[10pt, a4paper]{article}

\usepackage{lrec-coling2024} 
\usepackage{float}
\usepackage{natbib}
\usepackage{multibib}
\usepackage{inconsolata}
\usepackage{graphicx}
\usepackage{subfigure}
\usepackage{tabularx}
\usepackage{multirow}
\usepackage{algorithm,algcompatible,amsmath}
\usepackage{mathtools}
\usepackage{xcolor}
\usepackage{enumitem}
\usepackage{marvosym}
\usepackage{textcomp}
\usepackage{array}
\usepackage{booktabs}
\usepackage{hyperref}
\usepackage{url}
\usepackage{times,latexsym}
\usepackage{comment}
\usepackage{amssymb}
\usepackage{pifont}
\usepackage{bbm}
\usepackage{graphicx}
\usepackage{tabularx}
\usepackage{soul}
\usepackage{titlesec}

\title{MoELoRA: Contrastive Learning Guided Mixture of Experts on Parameter-Efficient Fine-Tuning for Large Language Models}

\name{\normalsize Tongxu Luo$^{1*}$\thanks{~~* Equal Contributions.}~
\quad Jiahe Lei$^{1*}$
\quad Fangyu Lei$^{1,2}$
\quad Weihao Liu$^{1}$ \\ 
\bf{Shizhu He$^{1,2}$
\quad Jun Zhao$^{1,2}$ 
\quad Kang Liu$^{1,2}$}
} 

\address{$^1$Institute of Automation, CAS \quad $^2$University of Chinese Academy of Sciences \\
\texttt{tongxuluo@163.com \quad \{shizhu.he, kliu\}@nlpr.ia.ac.cn}}

\abstract{
Fine-tuning is often necessary to enhance the adaptability of  Large Language Models (LLM) to downstream tasks. Nonetheless, the process of updating billions of parameters demands significant computational resources and training time, which poses a substantial obstacle to the widespread application of large-scale models in various scenarios. To address this issue, Parameter-Efficient Fine-Tuning (PEFT) has emerged as a prominent paradigm in recent research. However, current PEFT approaches that employ a limited set of global parameters (such as LoRA, which adds low-rank approximation matrices to all weights) face challenges in flexibly combining different computational modules in downstream tasks. In this work, we introduce a novel PEFT method: MoELoRA. We consider LoRA as Mixture of Experts (MoE), and to mitigate the random routing phenomenon observed in MoE, we propose the utilization of contrastive learning to encourage experts to learn distinct features. We conducted experiments on 11 tasks in math reasoning and common-sense reasoning benchmarks. With the same number of parameters, our approach outperforms LoRA significantly. In math reasoning, MoELoRA achieved an average performance that was 4.2\% higher than LoRA, and demonstrated competitive performance compared to the 175B GPT-3.5 on several benchmarks. 
 \\ \newline }

\begin{document}

\maketitleabstract

\begin{figure*}
\centering
\includegraphics[width=0.8\linewidth]{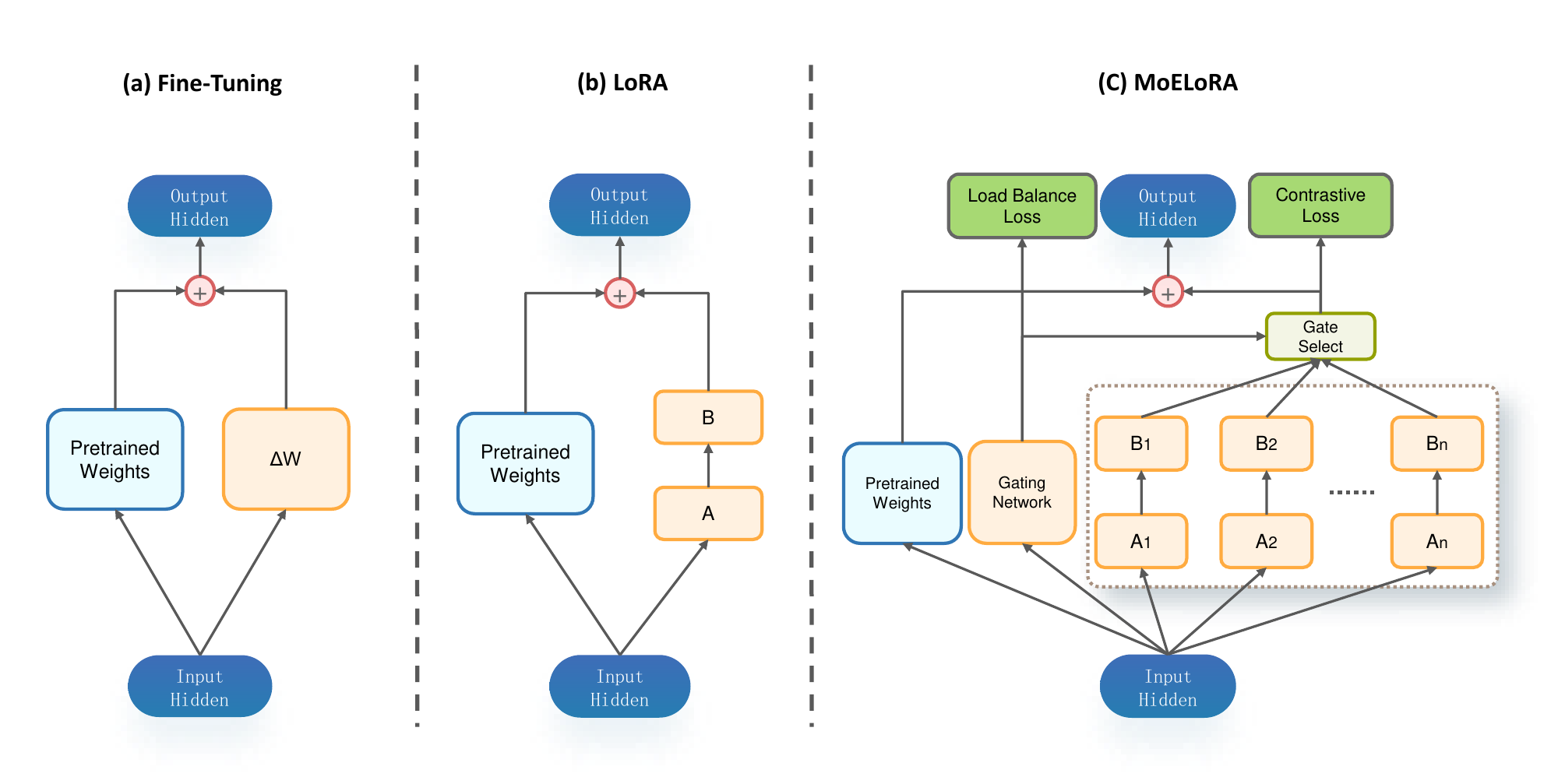}
\caption{The Different Architectures for (a)Fine-Tuning, (b)LoRA and (c)proposed method MoELoRA. $\Delta W$ denotes the gradient increment for the downstream tasks. LoRA decomposes $\Delta W$ into two matrices $A$ and $B$ and our proposed MoELoRA can select $A_i$ and $B_i$ corresponding to a specific task for better adaptation. In order to differentiate the capabilities of different experts, we employed contrastive learning on the outputs of the experts.
}
\label{moelora}
\end{figure*}

\section{Introduction}

With the rapid advancement of Large Language Models (LLMs) such as GPT3~\citep{brown2020language}, BLOOM~\citep{scao2022bloom} and LLaMA~\citep{touvron2023llama}, the successful application of self-supervised pretraining on unlabeled text data has presented unprecedented opportunities for enhancing downstream tasks. However, to fully harness the potential of these LLMs in practical applications, it is also necessary to continuously fine-tuning~\citep{wei2021finetuned,chung2022scaling} the LLMs based on the training data of specific tasks to meet the performance requirements of downstream tasks. 
The substantial number of parameters, often exceeding one billion, makes fine-tuning these LLMs a costly endeavor, demanding a significant investment in computational resources (Figure \ref{moelora}a). Therefore, in recent years, Parameter-Efficient Fine-Tuning (PEFT)~\citep{peft,zhang2023adaptive} techniques have emerged with the aim of reducing the cost of fine-tuning by freezing certain model weights or introducing smaller trainable modules.

In the continual exploration within this field, a series of methods such as LoRA ~\citep{hu2021lora}, AdaLoRA~\citep{zhang2023adaptive}, Adamix~\citep{wang2022adamix}, QLoRA~\citep{dettmers2023qlora} and LoRAHub~\citep{huang2023lorahub} have emerged, each offering unique perspectives on efficiently fine-tuning Large Language Models for better applicability in downstream tasks. LoRA (Figure \ref{moelora}b) introduces the concept of LoRA rank to reduce the number of trainable parameters. AdaLoRA builds upon LoRA's foundation, achieving a search-free approach that greatly simplifies the fine-tuning process. Adamix combines the MoE with Adapters to surpass the performance of LoRA. LoRAHub employs a gradient-free method~\citep{liu2020versatile} to perform weighted combinations of multiple LoRA weights, thereby better adapting to new downstream tasks.

However, current PEFT approaches that employ a limited set of global parameters face challenges in flexibly combining different computational modules in downstream tasks. Inspired by methods such as Mixture of Experts (MoE), Adamix, and LoRAHub, we propose a novel PEFT approach named MoELoRA. This method considers LoRA as a Mixture of Experts, leveraging the modeling capabilities of multiple experts for complex data domains, as well as utilizing LoRA's parameter-efficient characteristics. As well as Figure \ref{moelora}c, during both training and inference, only the LoRA selected by the gating network will be activated and only these "experts" relevant to specific tasks will participate in gradient updates or forward inference. However, applying MoE to LoRA presents challenges. Firstly, under the MoE architecture, gating network doesn't exhibit a preference for a particular expert, leading to a certain level of routing randomness~\citep{zuo2021taming}. Secondly, guiding experts to learn distinct features poses a challenging task.

To address these issues, we introduce contrastive learning among experts. Through this contrastive learning approach, we treat the outputs of the same expert as positive samples and the outputs of different experts as negative samples, encouraging experts to learn distinct features. In the end, we achieve performance surpassing LoRA under the same number of parameters. In math reasoning, MoELoRA averaged 4.2\% higher performance than LoRA, and in common-sense reasoning, it averaged 1.0\% higher than LoRA. Furthermore, MoELoRA exhibits competitive performance compared to the 175B GPT-3.5 on a few benchmarks.

In summary, our work makes the following contributions:

(1) We consider LoRA as Mixture of Experts and propose a novel PEFT method named MoELoRA, which leverages the MoE architecture to achieve dynamic combinations of multiple LoRA modules, better catering to the requirements of downstream tasks.

(2) In response to the random routing issue in using Mixture of Experts (MoE) for LoRA fusion, we propose employing contrastive learning to encourage experts to learn distinct features.


(3) We conduct experiments on 11 datasets for math reasoning and common-sense reasoning tasks, demonstrating that our approach outperforms LoRA in all tasks. The results of ablation experiments also show improvement in downstream tasks with contrastive learning. Furthermore, we perform tracking analysis of MoE routing to understand the impact of our method on the model's decision-making process.
\section{Related Work}
\subsection{Parameter-Efficient Fine-Tuning}
While fine-tuning with task-specific data sets, full-model fine-tuning not only demands substantial computational and storage resources but can also result in catastrophic forgetting. In contrast, Parameter-Efficient Fine-Tuning (PEFT)~\cite{peft} selectively adjusts a limited number of parameters or introduces additional trainable parameters rather than the entire backbone model, yet it still achieves comparable or even superior performance compared to full fine-tuning~\cite{ding2023parameter}. Prefix-tuning~\cite{li2021prefix} and Prompt-tuning~\cite{lester2021power} conditions frozen language models via trainable virtual token embeddings. Adapters~\cite{houlsby2019parameter,he2021towards,wang2022adamix} insert trainable adapter layers between existing layers in neural networks and fine-tune only them. \citet{hu2021lora} introduced LoRA, which using two low-rank matrices and exclusively fine-tuning LLMs. However, single LoRA cannot flexibly combine different computational modules in downstream tasks. We set up multiple LoRAs as distinct experts and dynamically combine them to achieve better PEFT.


\subsection{Mixture-of-Experts}
The Mixture of Experts (MoE) integrates the outputs of specialized sub-models, referred to as \textit{experts}, through an token-dependent ~\textit{router} mechanism. Assuming the existence of natural subsets in the dataset, such as originating from different domains or topics, a \textit{gating} network is employed to determine which expert should be trained. This enables each network to process a subset of the entire training dataset, addressing the challenge of generalization for a single model on complex datasets.

~\citet{shazeer2017outrageously} introduced the Sparsely Gated Mixture of Expert (MoE) models, employing a \textit{top-k} routing strategy to maintain sparsity while scaling the model parameters. This approach achieved a parameter scale of 137 billion in RNN-based networks, while ensuring low computational costs for both training and inference (e.g., FLOPs, parameters). By designing loss functions to enforce expert load balancing, this methodology resulted in state-of-the-art performance in language modeling and machine translation benchmarks.

Additionally, recent studies by GShard ~\citep{lepikhin2020gshard}, Switch-Transformer ~\citep{fedus2022switch}, BASELayer ~\citep{lewis2021base}, and Hash Layer ~\citep{roller2021hash} have focused on the development of large-scale Transformer-based models incorporating MoE, alongside the exploration of optimal training strategies to fully harness the model's capacity. In contrast to their work, we integrate MoE into PEFT and validate its effectiveness.

\subsection{Contrastive Learning}
Contrastive Learning ~\citep{hadsell2006dimensionality} has emerged as a powerful paradigm in the field of unsupervised representation learning. It aims to learn meaningful representations by maximizing the agreement between differently augmented views of the same data. Several studies ~\citep{zhuang2019local,misra2020self,chen2020simple} have introduced methods to align the representations of various augmentations applied to an image, leading to notable successes in computer vision. 


Contrastive learning has also proven to be a successful approach in NLP tasks. For instance, ~\citet{conneau2019unsupervised} introduced a contrastive learning framework tailored for acquiring multilingual representations, showcasing its efficacy in cross-lingual tasks. CERT ~\citep{fang2020cert} utilizes the method of back-translation to generate augmented versions of original sentences, while DeCLUTR ~\citep{giorgi2020declutr} posits that different segments within a document are similar to each other. CLEAR~\citep{wu2020clear}, adopts a structure with only an encoder, and acquire a noise-invariant sentence representation.

Furthermore, numerous variants and extensions of contrastive learning have been introduced to enhance its effectiveness. For example, ~\citet{chen2020simple} introduced SimCLR, which employs a set of data augmentations and a large batch size to achieve impressive results on various computer vision tasks. MoCo~\citep{he2020momentum} introduced a memory bank mechanism to enable more efficient contrastive learning. In this paper, we introduce the framework of contrastive learning into the MoE model, aiming to maximize the discrepancy in output distributions among different experts in order to capture diverse features in downstream tasks, mitigating the random routing phenomenon showed in ~\citet{zuo2021taming}.

\section{The Proposed Method}

\subsection{Framework of MoELoRA}\label{method}

MoELoRA combines the concept of MoE with LoRA, effectively increasing model parameters while maintaining the same computational cost to achieve superior performance. Specifically, our method is detailed as follows:

Firstly, we consider the traditional MoE architecture. For an input token $x \in \mathbb{R}^{d}$, we obtain the weight for each expert through a gating network $G:\mathbb{R}^{d}\mapsto \mathbb{R}^{n}$, resulting in $G(x) = [G(x)_1, G(x)_2, ..., G(x)_n]$, where $n$ represents the number of experts, and $G(x) \in \mathbb{R}^{n}$. Subsequently, we utilize these weights to linearly combine the outputs of different experts, yielding the output $y$ of the MoE layer:

\begin{equation}
y = \sum_{i = 1}^{n} G(x)_i \odot E_i(x)
\end{equation}

The essence of MoE lies in increasing the model's capacity while keeping the number of parameters for prediction and training constant. The gating network adopts a Top $k$ routing strategy, where only $k \ll n$ weights in $G(x)$ are non-zero. This means that despite adding more experts, which increases the overall model parameter count, only a small number of experts are involved in computations during both forward and backward passes, achieving sparsity.

Next, we consider the LoRA structure. Initially, the input $x$ undergoes a LoRA Dropout operation to enhance its generalization capability. Subsequently, it is projected downwards to $r$ ($r \ll d$) dimensions through $A(x)$, where $r$ represents the LoRA Rank. Following this, it is projected back up to $d$ dimensions through $B(x)$, and this process can be represented as:

\begin{equation}
A(x) = xA
\end{equation}
\begin{equation}
B(x) = xB
\end{equation}
\begin{equation}
LoRA(x) = B(A(x)) = xAB
\end{equation}

Where $A \in \mathbb{R}^{d \times r}$ and $B \in \mathbb{R}^{r \times d}$ are weight matrices.

We consider different LoRA modules as experts, forming the architecture of MoELoRA. For an input sample $x$, we first utilize the gating network to generate a weight vector $G(x)$. Subsequently, we apply these weights to different branches within each LoRA structure, resulting in multiple fine-tuned branches, denoted as $LoRA_i(x)$. Ultimately, we obtain the final MoELoRA prediction output by linearly combining these branches as follows:

\begin{equation}
MoELoRA(x) = \sum_{i = 1}^{n} G(x)_i \odot LoRA_i(x)
\end{equation}



\subsection{Challenge of MoELoRA}\label{challange}

\subsubsection{Load Imbalance}
Without intervention, Top $k$ MoE often assigns a large number of tokens to a few experts, while the remaining experts receive little or no tokens assigned~\citep{zuo2021taming}. This can lead to poor performance. Therefore, previous work~\citep{shazeer2017outrageously,fedus2022switch} used Load Balancing Loss to encourage balanced routing.

\subsubsection{Random Routing}
The MoE model exhibits a phenomenon, where the gating network shows no preference for any specific expert, resulting in a routing process that appears random. In such cases, due to the fact that each expert receives tokens generated by random routing ~\citep{zuo2021taming}, the content learned by all experts actually does not differ significantly. This contradicts the original intention of employing MoE, which is to break down a large problem into smaller subproblems, train different experts to address these subproblems effectively, and then combine the outputs of these experts. Therefore, addressing random routing presents a major challenge that must be overcome in the MoE architecture.

\begin{figure*}
\centering
\includegraphics[width=0.82\linewidth]{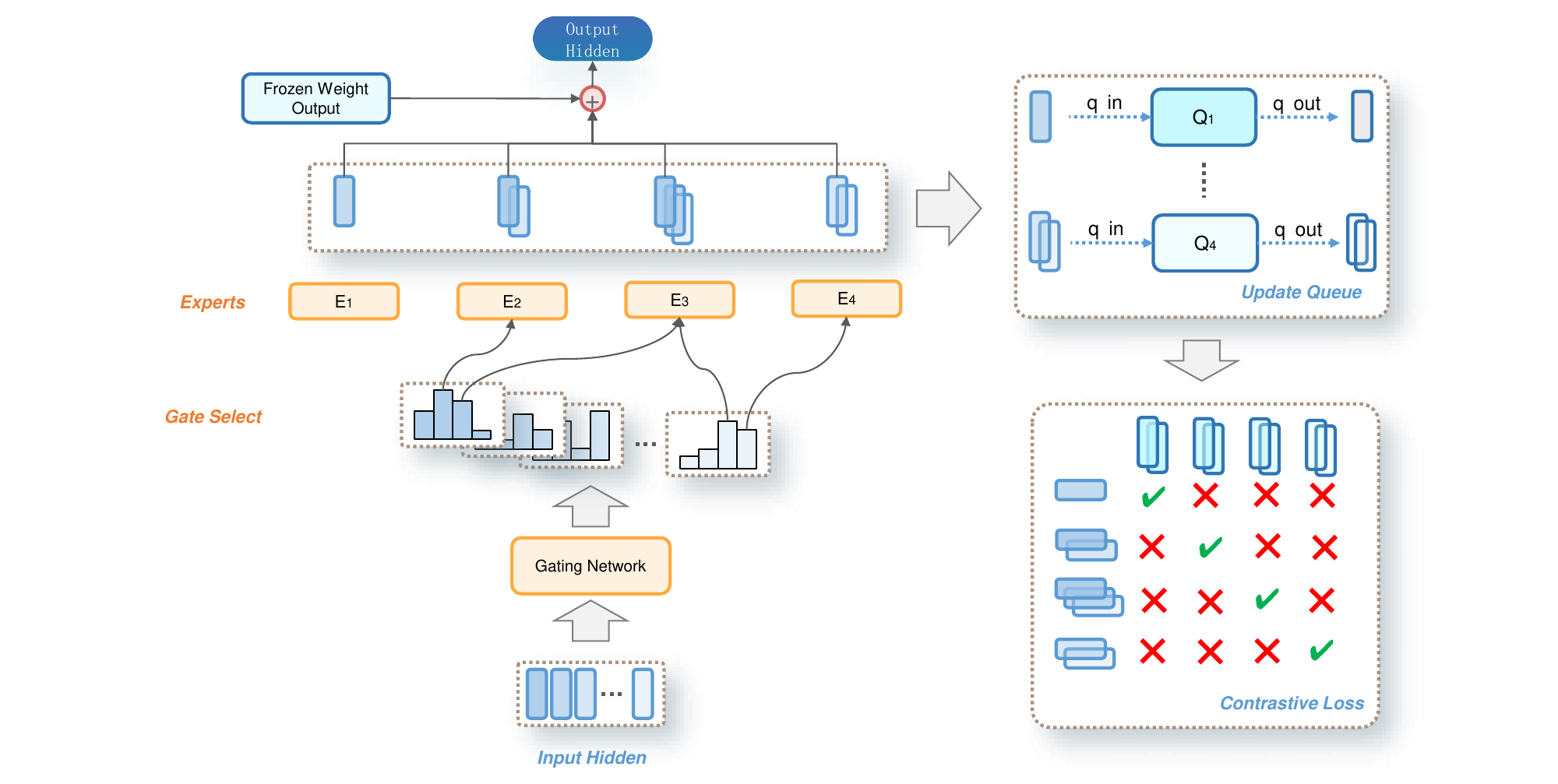}
\caption{As shown in the figure, it illustrates the process of calculating the Experts Contrastive Loss. The example uses a sentence input $h \in \mathbb{R}^{T \times d}$, where each token selects the top 2 experts. Initially, each expert updates its respective queue with tokens selected by that expert. Subsequently, the Contrastive Loss is computed using the samples from these queues.}\label{nceloss}
\end{figure*}

\subsection{Auxiliary loss} \label{loss}
\subsubsection{Load Balancing Loss}
During the training process, the gating network tends to converge towards a state wherein it consistently allocates substantial weights to a limited subset of experts ~\citep{zuo2021taming}, potentially resulting in an imbalanced distribution of workload among them. To address this concern, ~\citet{shazeer2017outrageously} and ~\citet{fedus2022switch} proposed the load-balancing loss and this paper, we adopt the latter.

Consider a training batch $B$ with $T$ tokens. Let $f_i$ represent the proportion of tokens assigned to the $i$-th expert, i.e.,

\begin{eqnarray}
f_i=\frac{1}{T}\sum_{x\in B}\mathbbm{1}\{\arg\max p(x)=i\}
\end{eqnarray}

Let $P_i$ be the average of all $T$ probabilities generated by the gating network for the $i$-th expert. $P_i$ can be expressed as:

\begin{eqnarray}
P_i=\frac{1}{T}\sum_{x\in B}p_i(x)
\end{eqnarray}

Based on the above equations, $f$ is non-differentiable while $P$ is differentiable. The Load Balancing Loss $\mathcal{L}_l$ is defined as the dot product between $f$ and $P$, making it differentiable, and it can be represented as:

\begin{eqnarray}
 \mathcal{L}_l= n\sum_{i=1}^{n}f_i(x)\cdot P_i
\end{eqnarray}

This loss optimizes "load balancing" from two perspectives: $f$ characterizes the distribution of the number of tokens assigned to each expert, while $P$ describes the distribution of the output from the gating network. When the gating network outputs an average probability distribution of $[1/n \cdots 1/n]$ for tokens in a batch, $\mathcal{L}_l$ achieves its minimum value, which is $n\sum_{i=1}^{n}1/n \cdot 1/n = 1$.

\subsubsection{Experts Contrastive Loss}
We introduce contrastive learning to encourage experts to learn different features and mitigate random routing. For each input token, we select the top $k$ experts using a gating network, ensuring that each token is assigned to some experts. To promote different experts in learning distinct content from the input $x \in \mathbb{R}^{T \times d}$ (where T represents the total number of token batches), an intuitive approach is as follows: For the $T_i$ tokens assigned to expert $E_i$, they should share a common attribute, for example, if $E_i$ specializes in processing "verb" type tokens, then the common attribute among the tokens assigned to this expert is "verbs." For these "verb" type tokens, after being processed by $E_i$, they should be sufficiently close in the semantic space. Conversely, for two experts $E_i$ and $E_j$, since we expect them to learn different features, the tokens they process should be far apart in the semantic space. This can be expressed simply as:

\begin{eqnarray}
d(E_i(x_k), E_i(x_m)) \ll d(E_i(x_k), E_j(x_n))
\end{eqnarray}

Therefore, we can employ a contrastive learning approach proposed in ~\citet{he2020momentum}, where the outputs of the same expert are treated as positive samples, while the outputs of different experts are considered negative samples. Given input $x \in \mathbb{R}^{T \times d}$, the expert model outputs $E(x) = [E_1(x), E_2(x), \cdots, E_n(x)]$, where $E_i(x) \in \mathbb{R}^{t_i \times h}$, and $t_i$ represents the number of tokens activated by the $i$-th expert, satisfying the relationship $T \cdot top\ k = \sum t_i$. As per our definition of positive and negative samples in expert contrastive learning, let $q \in E_i(x)$ and $k_+ \in E_i(x)$.

Ultimately, for the $i$-th expert, the Experts Contrastive Loss can be defined as:

\begin{eqnarray}
\mathcal{L}_{E_i} = - \sum_{q\ne k_+} log \frac{exp(q \cdot k_+ / \tau )}{\sum_{k \in E(x)}exp(q \cdot k / \tau)}
\end{eqnarray}

Here, $\tau$ represents the temperature coefficient, controlling the distribution shape of $q \cdot k$. When $\tau$ increases, it smoothens the distribution of $q \cdot k$, reducing the discriminative power of $\mathcal{L}_E$ over all negative samples. Conversely, a lower $\tau$ value makes the model focus more on the negative samples during training. In Figure \ref{nceloss}, we illustrate the detailed calculation process of the Experts Contrastive Loss.

Finally, the Auxiliary Loss we adopt is defined as:

\begin{eqnarray}
\mathcal{L} = \alpha \cdot \mathcal{L}_{l} + \beta \cdot \mathcal{L}_{E}
\end{eqnarray}

where $\alpha$ and $\beta$ are hyperparameters.

\section{Experiments}\label{experiments}
\subsection{Experimental Setup}

\begin{table*}
\centering
\begin{tabular}{ll|ccccc}
\hline
\textbf{Task} & \textbf{Metric} & \textbf{Series-Adapter} & \textbf{Parallel-Adapter} & \textbf{LoRA} & \textbf{MoELoRA} & \textcolor{gray}{GPT-3.5} \\
\hline
AddSub & Acc & 69.6 & 77.2 & 84.8 & \textbf{88.6} & \textcolor{gray}{85.3} \\
AQuA & Acc & 15.6 & 9.8 & 17.6 & \textbf{25.5} & \textcolor{gray}{38.9}  \\
gsm8k & Acc & 18.5 & 22.7 & 31.1 & \textbf{32.6} & \textcolor{gray}{56.4} \\
MultiArith & Acc & 88.3 & 83.3 & 88.3 & \textbf{95.0} & \textcolor{gray}{83.8} \\
SingleEQ & Acc & 79.4 & 81.3 & 90.2 & \textbf{94.1} & \textcolor{gray}{88.1} \\
SVAMP & Acc & 52.0 & 57.0 & 65.0 & \textbf{66.0} & \textcolor{gray}{69.9} \\
\hline
\textbf{Avg.} & & 53.9 & 55.2 & 62.8 & \textbf{67.0} & \textcolor{gray}{70.4} \\
\textbf{Param.} & & 200M & 200M & 18.9M & 18.9M & \textcolor{gray}{175B} \\
\hline
\end{tabular}
\caption{Results on math reasoning tasks, "Param." represents the number of trainable parameters. \textit{Series-Adapter, Parallel-Adapter and GPT-3.5} results are taken from ~\citet{hu2023llm}}\label{table:math_result}
\end{table*}

\subsubsection{Dataset}
We evaluated LoRA and MoELoRA and other adapters on math reasoning and common-sense reasoning tasks. Our math reasoning dataset, as well as all the rationales for the samples, are taken from ~\citet{hu2023llm}. All rationales for the samples are generated through zero-shot-CoT~\citep{kojima2022large} on GPT-3.5, but without undergoing any error filtering. The math reasoning tasks includes a total of 6 benchmarks: AddSub~\citep{hosseini2014learning}, AQuA~\citep{ling2017program}, gsm8k~\citep{cobbe2021training}, MultiArith~\citep{roy2016solving}, SingleEQ~\citep{koncel2015parsing}, and SVAMP~\citep{patel2021nlp}. The Common-sense tasks we selected includes 5 benchmarks: namely ARC-C, ARC-E~\citep{chollet2019measure}, BoolQ~\citep{clark2019boolq}, OBQA~\citep{mihaylov2018can}, and PIQA~\citep{bisk2020piqa}.

\subsubsection{Implementation Details}
We using the LLaMA-7b ~\citep{touvron2023llama} as the Large Language Model. We conducted a comparison between Series-Adapter, Parallel-Adapter, LoRA and MoELoRA. We introduce LoRA or MoELoRA into the 'q\_proj' and 'p\_proj' of LLaMA. We set LoRA and MoELoRA with the same number of trainable parameters, demonstrating that MoELoRA outperforms LoRA significantly under the same settings. Subsequently, we conducted ablation experiments to analyze the various design components of MoELoRA.

In experiments, as AdapterH ~\citep{houlsby2019parameter} and AdapterP ~\citep{pfeiffer2020mad} are Series adapters, and AdapterP outperforms AdapterH, we use AdapterP with bottleneck size 768 as Series Adapter. For Parallel-Adapter ~\citep{pfeiffer2020mad}, the adapter layers have been placed in multi-head attention modules with a bottleneck size of 256. For LoRA, we set the LoRA Rank to $R=36$, while for MoELoRA, we set the LoRA Rank to $R=32$, with a total of $n=8$ experts, each having a LoRA Rank of $r=4$. This configuration ensured that LoRA and MoELoRA had an equal number of trainable parameters. For loss, $\tau$ is set to 0.07. $\alpha$ and $\beta$ are set to 0.01. All our experiments were conducted on a single RTX3090.

\subsection{Main Results}


\begin{table*}
\centering
\begin{tabular}{ll|cccccc}
\hline
\textbf{Task} & \textbf{Metric} & \textbf{Series-Adapter} & \textbf{Parallel-Adapter} & \textbf{LoRA} & \textbf{MoELoRA} & \textcolor{gray}{GPT-3.5} \\
\hline
ARC-C & Acc & 57.1 & 57.3 & 70.5 & \textbf{72.2} & \textcolor{gray}{79.9} \\
ARC-E & Acc & 74.5 & 73.7 & 85.3 & \textbf{85.6} & \textcolor{gray}{89.8} \\
BoolQ & Acc & 63.0 & 67.9 & 72.6 & \textbf{73.7} & \textcolor{gray}{73.1} \\
OBQA & Acc & 72.4 & 75.2 & 82.2 & \textbf{83.6} & \textcolor{gray}{74.8} \\
PIQA & Acc & 79.2 & 76.4 & 84.7 & \textbf{85.6} & \textcolor{gray}{85.4} \\
\hline
\textbf{Avg.} & & 69.2 & 70.1 & 79.1 & \textbf{80.1} & \textcolor{gray}{80.6} \\
\textbf{Param.} & & 200M & 200M & 18.9M & 18.9M & \textcolor{gray}{175B} \\
\hline
\end{tabular}
\caption{Results for mixed training on common-sense reasoning tasks, "Param." represents the number of trainable parameters.}\label{table:common_result}
\end{table*}

Table \ref{table:math_result} presents the performance on six math reasoning tasks benchmarks. In the AddSub, MoELoRA achieved a higher accuracy compared to LoRA by 3.8, and it also outperformed GPT-3.5 by 3.3 points. In the case of AQuA, MoELoRA showed an accuracy improvement of 7.9 over LoRA. For the gsm8k, MoELoRA's accuracy exceeded LoRA by 1.5. In the MultiArit, MoELoRA demonstrated an accuracy increase of 6.7 compared to LoRA, and it also outperformed GPT-3.5 by 11.2. In SingleEQ, MoELoRA's accuracy was 3.9 higher than LoRA, and it surpassed GPT-3.5 by 6.0. Finally, in the SVAMP, MoELoRA achieved a 1.0 accuracy improvement over LoRA. Our experiments have demonstrated that, with the same number of parameters, MoELoRA consistently outperforms LoRA in all aspects. On average accuracy, MoELoRA exhibits a 4.2 improvement over LoRA, surpassing the baseline LoRA comprehensively. Furthermore, MoELoRA remains highly competitive even when compared to GPT-3.5, which has nearly $10^4$ times more parameters.


Table \ref{table:common_result} showcases the performance of LoRA, MoELoRA, and GPT-3.5 on five common-sense reasoning benchmarks. In ARC-C, MoELoRA achieved an accuracy 1.7 higher than LoRA. In ARC-E, MoELoRA's accuracy was 0.3 higher than LoRA. For BoolQ, MoELoRA surpassed LoRA by 1.1 and also outperformed GPT-3.5 by 0.6. On OBQA, MoELoRA's accuracy exceeded LoRA by 1.4 and GPT-3.5 by 8.8. In the case of PIQA, MoELoRA's accuracy was 0.9 higher than LoRA and 0.2 higher than GPT-3.5. Our experiments have demonstrated that, with the same number of parameters, MoELoRA exhibits a 1.0\% improvement over LoRA on the common-sense reasoning tasks, and it remains competitive compared to GPT-3.5 on a few benchmarks.

\begin{table}[H]
\centering
\begin{tabular}{l|cc}
\hline
\textbf{Task} & \textbf{Contrastive Loss(w/o)} & \textbf{Total}\\
\hline
AddSub & 86.1 & \textbf{88.6} \\
AQuA & 21.6 & \textbf{25.5} \\
gsm8k & 30.7 & \textbf{32.6} \\
MultiArith & 91.7 & \textbf{95.0} \\
SingleEQ & 88.2 & \textbf{94.1} \\
SVAMP & 65.5 & \textbf{66.0}\\
\hline
Avg. & 64.0 & \textbf{67.0} \\
\hline
\end{tabular}
\caption{Results of ablation experiments on the Experts Contrastive Loss in the math reasoning tasks.}\label{table:loss_math}
\end{table}

\begin{table}[H]
\centering
\begin{tabular}{l|cc}
\hline
\textbf{Task} & \textbf{Contrastive Loss(w/o)} & \textbf{Total}\\
\hline
ARC-C & 71.4 & \textbf{72.2} \\ 
ARC-E & 85.3 & \textbf{85.6} \\    
BoolQ & 73.6 & \textbf{73.7} \\
OBQA & 81.8 & \textbf{83.6} \\
PIQA & 83.9 & \textbf{85.6} \\
\hline
Avg. & 79.2 & \textbf{80.1} \\
\hline
\end{tabular}
\caption{Results of ablation experiments on the Experts Contrastive Loss in the common-sense reasoning tasks.}\label{table:loss_common}
\end{table}

\subsection{Ablation Studies}
\subsubsection{Ablations on Auxiliary Loss}

To validate the effectiveness of our Experts Contrastive Loss, we conducted ablation experiments. Tables \ref{table:loss_math} display the results of ablation experiments on math reasoning tasks, and Table \ref{table:loss_common} presents the results for common-sense reasoning tasks. In these experiments, we kept LoRA Rank at $R=32$, with a total of $n=8$ experts, and utilized the setting where each token is assigned to the top 2 activated experts. The experimental outcomes indicate that removing the expert contrastive loss results in an average decrease of 3.0 in math reasoning tasks and an average decrease of 0.9 in common-sense reasoning tasks. These experiments provide evidence of the significant improvement in performance attributed to the Experts Contrastive Loss in MoELoRA.

\subsubsection{Ablations on Selecting Top-k per Token}

\begin{table}[H]
\centering
\begin{tabular}{l|ccc}
\hline
\textbf{Task} & \textbf{Top-1} & \textbf{Top-2} & \textbf{Top-4}\\
\hline
AddSub & 86.1 & \textbf{88.6} & 87.3 \\
AQuA & 23.5 & \textbf{25.5} & 19.6\\
gsm8k & 32.2 & \textbf{32.6} & 27.7\\
MultiArith & 89.2 & \textbf{95.0} & 93.3\\
SingleEQ & 91.2 & \textbf{94.1} & 89.2\\
SVAMP & 65.5 & \textbf{66.0} & \textbf{66.0}\\
\hline
Avg. & 64.6 & \textbf{67.0} & 63.9\\
\hline
\end{tabular}
\caption{Results of ablation experiments on Selecting Top-n per Token in the math reasoning tasks.}\label{table:topn_math}
\end{table}

\begin{table}[H]
\centering
\begin{tabular}{l|ccc}
\hline
\textbf{Task} & \textbf{Top-1} & \textbf{Top-2} & \textbf{Top-4}\\
\hline
ARC-C & 69.1 & \textbf{72.2} & 71.2\\ 
ARC-E & 84.1 & \textbf{85.6} & 85.0\\    
BoolQ & 72.6 & 73.7 & \textbf{74.0}\\
OBQA & 82.4 & \textbf{83.6} & 81.2\\
PIQA & 83.6 & \textbf{85.6} & 84.2\\
\hline
Avg. & 78.4 & \textbf{80.1} & 79.1\\
\hline
\end{tabular}
\caption{Results of ablation experiments on Selecting Top-n per Token in the common-sense reasoning tasks.}\label{table:topn_common}
\end{table}

Simultaneously, we conducted experiments involving the selection of the top-k experts for each token. We fixed LoRA Rank at $R = 32$ and employed a total of $n = 8$ experts. Surprisingly, we found that the performance exhibited a significant improvement when using the top-2 experts, as compared to top-1 and top-4 experts.Table \ref{table:topn_math} displays the results for math reasoning tasks, and Table \ref{table:topn_common} presents the outcomes for common-sense reasoning tasks.

\section{Analysis}\label{analysis}

\subsection{Why The Improvement In Common-sense Tasks Is So Inconspicuous?}\label{bad_common_sense}

In Appendix \ref{appendix:A}, we displays four different formats of benchmarks for common-sense tasks, namely ARC, BoolQ, OBQA, and PIQA. Each of these benchmarks necessitates that the LLM possesses corresponding knowledge, which places significant demands on the LLM's pretraining efficacy. Furthermore, during fine-tuning, if knowledge cannot be effectively injected, then fine-tuning on common-sense tasks becomes futile. Therefore, the performance on common-sense tasks relies more on the reservoir of knowledge that LLMs have accumulated during the pretraining phase. While PEFT does have an impact on common-sense tasks, it ultimately cannot address the issue that LLMs may lack relevant knowledge.

~\citet{geva2020transformer,dai2021knowledge} have demonstrated that Feedforward Networks (FFNs) can be interpreted as memory networks capable of storing substantial amounts of knowledge. In moefication, ~\citet{zhang2022moefication} analyzed the activation patterns of FFNs within Transformer models and discovered a phenomenon wherein only a small fraction of neurons are activated for a single input. Their findings corroborate that a Transformer model can be transformed into an equivalent Mixture-of-Experts (MoE) model.

\subsection{Tracing tokens through Experts}
\begin{figure}[H]
\centering
\includegraphics[width=\linewidth]{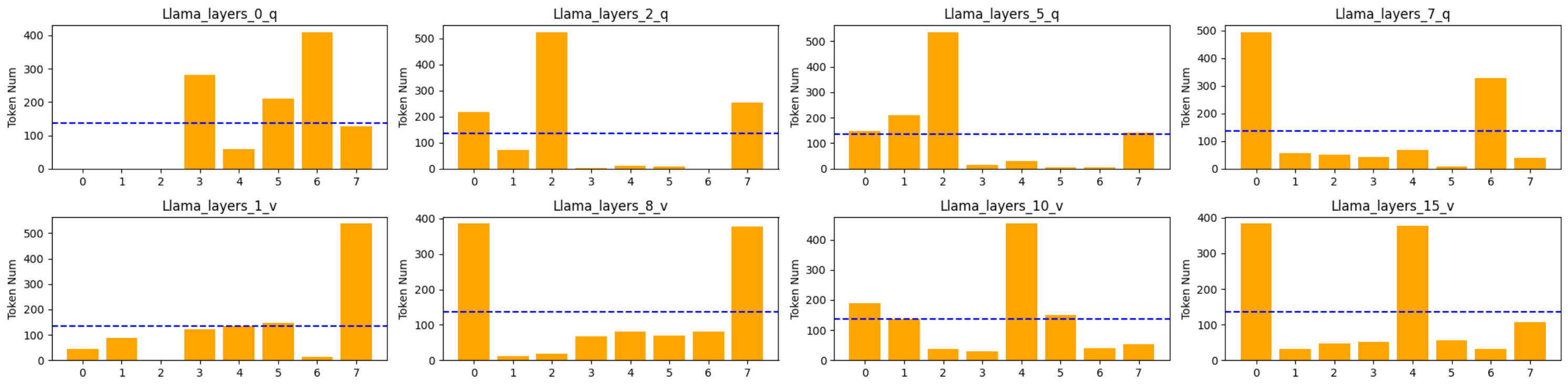}
\caption{The figure displays the routing of all numeric tokens, which are often assigned to specific experts.}
\label{num_routing}
\end{figure}

In our math reasoning task, we tracked the token routing within the MoE to analyze whether the phenomenon of random routing has been mitigated. Firstly, we traced the routing of all numerical tokens in certain layers, as shown in Figure \ref{num_routing}. We observed that there are always a few specific experts who excel at handling numerical tokens.

\begin{figure}[H]
\centering
\includegraphics[width=\linewidth]{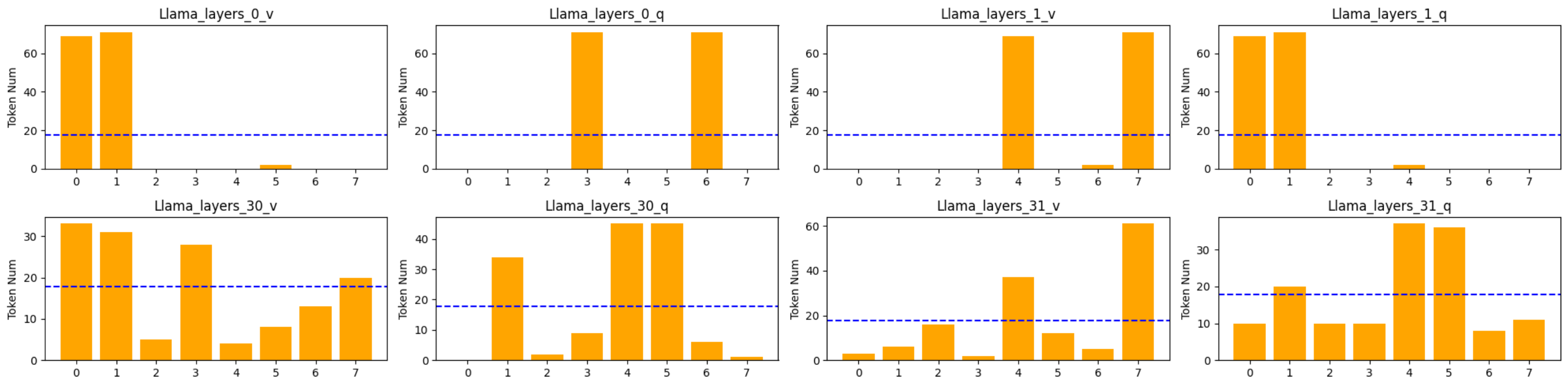}
\caption{The figure displays the routing of numerical token '2' .}
\label{token_2}
\end{figure}

\begin{figure}[H]
\centering
\includegraphics[width=\linewidth]{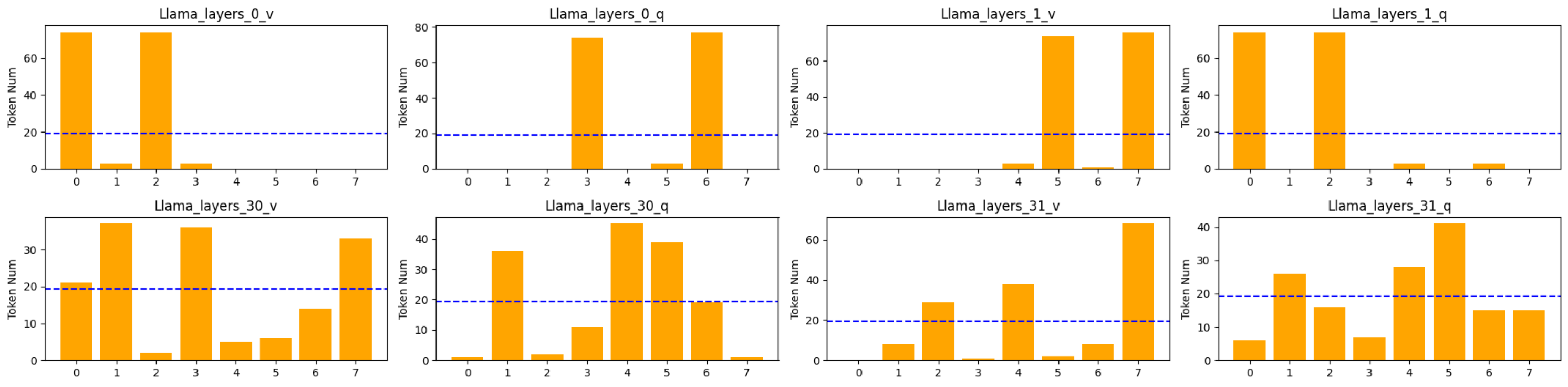}
\caption{The figure displays the routing of numerical token '4' .}
\label{token_4}
\end{figure}

We also observed that for specific numerical tokens, such as '2' or '4' in figure \ref{token_2} and \ref{token_4}, they are routed to specific experts for processing in the early layers. However, due to the influence of the attention mechanism, as the layers progress and tokens assimilate a wealth of information, their routing becomes more uniform.

Furthermore, to our surprise, we found that the load is not particularly balanced. However, upon closer examination, this is expected because the dataset inherently contains variations in token frequency. Some tokens appear more frequently in the dataset, while others occur less often. See table \ref{Frequency}. The differing occurrence frequencies of tokens in the dataset make achieving load balance a challenging task. But Load balancing Loss is still needed, otherwise some experts will not be assigned tokens from beginning to end.

\section{Conclusions and Future Work}
We have introduced a novel Parameter-Efficient Fine-Tuning method called MoELoRA and mitigate the random routing phenomenon observed in
MoE through contrastive learning. Additionally, we conducted extensive experiments on 11 math reasoning and common-sense reasoning datasets. In math reasoning, MoELoRA averaged 4.2\% higher performance than LoRA, and in common-sense reasoning, it averaged 1.0\% higher than LoRA. The results demonstrate that MoELoRA consistently outperforms LoRA across all tasks. Furthermore, when compared to the GPT-3.5 model, MoELoRA demonstrates its competitive performance.

\textbf{Future Work:} In Section \ref{bad_common_sense}, we mentioned the limited improvements on common-sense tasks. Therefore, it may be worthwhile to explore MoELoRA by reframing common-sense tasks as knowledge editing tasks. In addition,  we can potentially adopt LoRA modules trained on different tasks for each expert, freeze them, and only train the gating network.

\begin{table}[H]
\centering
\begin{tabular}{l|cc}
\hline
& \textbf{Token} & \textbf{Frequency}\\
\hline
top-1 & \textunderscore & 748 \\
top-2 & . & 421 \\
top-3 & \textunderscore the & 371 \\
... & ... & ... \\
top-51 & \textunderscore her &  46 \\
top-68 & \textunderscore game &  24 \\
top-100 & \textunderscore scored &  14 \\
\hline
\end{tabular}
\caption{On the development set of the MultiArith benchmark, we have conducted a statistical analysis of token frequencies, specifically focusing on the fast decay of the token frequency.}\label{Frequency}
\end{table}

\nocite{*}
\section{References}\label{sec:reference}

\bibliographystyle{lrec-coling2024-natbib}
\bibliography{lrec-coling2024-example}


\appendix
\section{Case Study on Common-sense Tasks}\label{appendix:A}

\begin{table*}
\centering
\begin{tabular}{l|p{8cm}|p{4cm}}
\hline
\textbf{Benchmark} & \textbf{Question} & \textbf{Answer}\\
\hline
ARC & George wants to warm his hands quickly by rubbing them. Which skin surface will produce the most heat? & The correct answer is answer1. \\
& Answer1: dry palms & \\
& Answer2: wet palms & \\
& Answer3: palms covered with oil & \\
& Answer4: palms covered with lotion & \\
\hline
BoolQ & Do iran and afghanistan speak the same language? True or False. & The correct answer is true. \\
\hline
OBQA & The sun is responsible for & The correct answer is answer4. \\
& Answer1: puppies earning new tricks & \\
& Answer2: children growing up and getting old & \\
& Answer3: flowers wilting in a vase & \\
& Answer4: plants sprouting, blooming and wilting & \\
\hline
PIQA & When boiling butter,when it's ready, you can & The correct answer is solution2. \\
& Solution1: Pour it onto a plate & \\
& Solution2: Pour it into a jar & \\
\hline
\end{tabular}
\caption{Case study displays four different formats of benchmarks for the common-Sense tasks, each of which essentially requires the LLM to possess relevant knowledge.}\label{case}
\end{table*}

To investigate why the improvement on common-sense tasks is relatively small, we conducted a case study on several benchmarks for this task, as listed in Table \ref{case}.

\end{document}